# Developing Federated Time-to-Event Scores Using Heterogeneous Real-World Survival Data

Siqi Li[1#], Yuqing Shang[1#], Ziwen Wang[1], Qiming Wu[1], Chuan Hong[2], Yilin Ning[1], Di Miao[1], Marcus Eng Hock Ong[3,4,5], Bibhas Chakraborty[1,2,3,6], Nan Liu[1,3,7]*

[1] Centre for Quantitative Medicine, Duke-NUS Medical School, Singapore, Singapore

[2] Department of Biostatistics and Bioinformatics, Duke University, Durham, NC, USA

[3] Programme in Health Services and Systems Research, Duke-NUS Medical School, Singapore, Singapore

[4] Health Services Research Centre, Singapore Health Services, Singapore, Singapore

[5] Department of Emergency Medicine, Singapore General Hospital, Singapore, Singapore

[6] Department of Statistics and Data Science, National University of Singapore, Singapore, Singapore

[7] Institute of Data Science, National University of Singapore, Singapore, Singapore

[#] Co-first author

* Correspondence: Nan Liu, Centre for Quantitative Medicine, Duke-NUS Medical School, 8 College Road, Singapore 169857, Singapore. Phone: +65 6601 6503. Email: liu.nan@duke-nus.edu.sg

## Abstract

**Objective**

Survival analysis serves as a fundamental component in numerous healthcare applications, where the determination of the time to specific events (such as the onset of a certain disease or death) for patients is crucial for clinical decision-making. Scoring systems are widely used for swift and efficient risk prediction. However, existing methods for constructing survival scores presume that data originates from a single source, posing privacy challenges in collaborations with multiple data owners.

**Materials and Methods**

We propose a novel framework for building federated scoring systems for multi-site survival outcomes, ensuring both privacy and communication efficiency. We applied our approach to sites with heterogeneous survival data originating from emergency departments in Singapore and the United States. Additionally, we independently developed local scores at each site.

**Results**

In testing datasets from each participant site, our proposed federated scoring system consistently outperformed all local models, evidenced by higher integrated area under the receiver operating characteristic curve (iAUC) values, with a maximum improvement of 11.6%. Additionally, the federated score's time-dependent AUC(t) values showed advantages over local scores, exhibiting narrower confidence intervals (CIs) across most time points.

**Discussion**

The model developed through our proposed method exhibits effective performance on each local site, signifying noteworthy implications for healthcare research. Sites participating in our proposed federated scoring model training gained benefits by acquiring survival models with enhanced prediction accuracy and efficiency.

**Conclusion**

This study demonstrates the effectiveness of our privacy-preserving federated survival score generation framework and its applicability to real-world heterogeneous survival data.

# 1. Background and Significance

Risk prediction models play an essential role in guiding clinical decision-making, with interpretability being a key factor for healthcare practitioners considering their integration into routine clinical practice. Scoring systems, recognized for their interpretability[1], have gained widespread acceptance in healthcare, proving their utility across diverse diagnostic areas in medicine[2]. Unlike more complex computational approaches, scoring systems offer a unique advantage by enabling rapid and straightforward risk assessments for critical medical conditions, relying on simple arithmetic operations involving only a few numbers[1]. While traditional scoring systems, such as the Glasgow Coma Scale[3], are rooted in clinicians' domain knowledge, there has been a significant move toward adopting more data-driven approaches in recent years[4–7]. This shift reflects an evolving landscape where empirical data and analytics are increasingly harnessed to enhance the accuracy and reliability of risk assessments in healthcare.

Beyond their diverse applications in healthcare, a large number of scoring systems have primarily been designed to assess outcomes at a single time point[8–10]. Generally, medical prognostic models can be categorized into two classes: predicting outcomes at a single time point or predicting time-to-event outcomes[11]. While it is common for medical research to simplify outcomes such as mortality into binary variables for quick and efficient risk assessment[12], this approach may not adequately address research questions focused on the duration until an event occurs, such as estimating survival rates over time post-treatment or estimating the probability of surviving beyond a prespecified time interval[13]. For example, in a case-control study[14] investigating biomarkers for lethal prostate cancer, baseline logistic regressions were found to underutilize the available data. This was demonstrated by their reduced predictive power and increased standard errors in real data applications when compared to survival analysis[14]. This highlights the limitations of traditional scoring systems in capturing the complexities in time-to-event data, underscoring the need for more sophisticated analytical methods in certain scenarios.

Although time-to-event clinical scores have been previously developed[15–17] using existing score development tools[18,19], these advancements have predominantly relied on single-source data. In recent years, a growing trend in cross-institutional collaborations has emerged to expedite medical research [20] and provide validated quality assessments of clinical prediction models[21]. Like other clinical prediction models, the ideal validation of scoring systems would involve the use of centralized pooled data[22]. However, accomplishing this can be challenging due to various privacy constraints[23,24] related to data sharing. To overcome such barriers, federated learning (FL) has emerged as a valuable approach. FL enables multi-site healthcare collaborations by safeguarding privacy through the collective training of algorithms. The approach avoids the need to exchange patient-level data[25], as model training is distributed to data owners, and their results are aggregated[26].

## 2. Objective

In this work, we bridge the gap by presenting a federated scoring system framework designed for handling multi-site time-to-event clinical outcomes. Our method prioritizes communication efficiency and user-friendly implementation, relying solely on the broadcast and reception of intermediate summary statistics, without the exchange of sensitive patient-level information. To validate the effectiveness of our method, we conducted a proof-of-concept experiment using two real-world heterogeneous emergency department (ED) datasets from Singapore and the United States. The experiment demonstrates that our method could bring participating sites benefits that cannot be achieved via local survival analyses.

## 3. Methods

The integration of scoring systems and other models in medical science often prioritizes a model's degree of parsimony, meaning that the model should be sparse, use the fewest variables necessary, and also possess strong predictive accuracy[27]. This aligns with the general expectation in implementing risk stratification scoring systems for simple, quick, yet highly accurate risk assessment in clinical practice. In our study, we emphasize the importance of ensuring the final federated survival scoring system generated by our proposed framework maintains this level of parsimony while also affording clinicians the flexibility to incorporate their domain expertise into the model selection and refinement process.

### 3.1 The proposed FedScore-Surv Framework

We propose the FedScore-Surv framework for survival outcome, consisting of five modules: 1) federated variable ranking; 2) federated variable transformation; 3) federated score derivation; 4) federated model selection and 5) federated model evaluation. The workflow of FedScore-Surv is illustrated in Figure 1.

**1) Federated Variable Ranking**

To achieve a good control of model parsimony, we first conduct variable selection to pre-identify a set of unified candidate variables across all participating sites. We recommend users to check for multicollinearity among the candidate variables first and remove variables if needed to obtain a more reliable feature importance analysis[28]. We employ random survival forests[29] for variable importance measurement, a well-established approach widely applied[18,30,31] in clinical science. The variable ranking is first performed independently at each local site, and then a global variable ranking is obtained by weighting the local ranks across all $K$ sites. Specifically, for a single variable $X_m$ where $1 \leq m \leq P$ and $P$ is the total number of predictors, let integer $q_j \in N$ denote its rank at site $j$. The variable's global ranking is obtained by mapping all values of $\sum_{j=1}^{K} w_j q_j$ for each site to the integer set $[1, P] \subset Z$. Here, remaining the same throughout this manuscript, $w_j$ is defined as the normalized weight for site $j$ that satisfies $\sum_{j=1}^{K} w_j = 1$. The default setting for the weight is $w_j = S_j/S_0$, where $S_j$ is the sample size of site $j$, and $S_0$ is the total sample size. Users may also define their own weights to accommodate their specific research considerations.

**2) Federated Variable Transformation**

We next transform continuous variables into categorical variables, which is a common strategy[32–38] in the development of clinical scoring systems for modelling nonlinear effects[1,7]. In our study, the default number of categories of a given continuous variable is set to be five, and the quantiles are set correspondingly at $0\%, k_1\%, k_2\%, k_3\%, k_4\%$, and $100\%$, where the default value of $k_1, k_2, k_3, k_4$ are 20, 40, 60 and 80, respectively. Categories of a given variable may be combined if the maximum is surpassed or two neighbouring categories have similar effect sizes in the proceeding modelling steps. A unified cutoff for each continuous variable is then calculated by weighting the $k$ values acquired at each site using the same weight $\sum_{j=1}^{K} w_j = 1$ as defined previously for global ranking.

**3) Federated Score Derivation**

We use Cox regression[39] to model time-to-event outcomes and obtain scores. The Cox model assumes the hazard at time $t$ given the vector of $p$-dimensional predictors $X$ follows:

$$\lambda(t|X) = \lambda_0(t)\exp(\boldsymbol{\beta}^T X)$$

Here, $\boldsymbol{\beta}$ is the regression coefficient vector and $\lambda_0(t)$ is the baseline hazard function.

Federated regression models can be realized through a variety of existing FL frameworks, including both engineering-based[40], model-agnostic FL frameworks like FedAvg[41] that requires multiple iterations, and statistics-based[40], model specific FL techniques[42–47] that necessitate only one or a few rounds of communication. For easy demonstration purpose, we employed a communication-efficient, privacy preserving distributed algorithm called ODACH[48] to perform federated Cox regression, which requires only one round of communication and can be easily employed regardless of the existence of clients' data protection systems, since it only requires participants to receive and broadcast non-sensitive information without the need for a server. In particular, ODACH allows for cross-site data heterogeneity and has been demonstrated to have low bias and high statistical efficiency using simulation studies[48].

Let $\{T_{ij}, \delta_{ij}, x_{ij}\}$ represent the observation of the $i$-th patient at the $j$-th site, where $T_{ij}$ is the observed survival time, $\delta_{ij}$ equals 0 indicating censoring and equals 1 indicating an event, and $x_{ij}$ is a p-dimensional covariate vector. In an ideal scenario where data can be pooled, the pooled estimator of Cox regression can be obtained by maximizing the global log partial likelihood function, written as: $L(\boldsymbol{\beta}) = \frac{1}{N}\sum_{j=1}^{K}\sum_{i=1}^{n_j} \delta_{ij} \log \frac{\exp(\boldsymbol{\beta}^T x_{ij})}{\sum_{s \in R_j(T_{ij})} \exp(\boldsymbol{\beta}^T x_{ij})}$. Here, $R_j(t) = \{i; T_{ij} \geq t\}$ denotes the risk set at time $t$ for site $j$. When there is privacy restriction and data cannot be shared across sites, we employ ODACH to utilize information from each local site (where data is accessible) with the first-order and second-order gradients of the likelihood function from remote sites (where data is not accessible) to construct an approximation of the global log partial likelihood function. In particular, the surrogate likelihood function[44] obtained at site $j$ is

$$\tilde{L}_j(\boldsymbol{\beta}) = L_j(\boldsymbol{\beta}) + \langle \nabla L(\overline{\boldsymbol{\beta}}) - \nabla L_j(\overline{\boldsymbol{\beta}}), \boldsymbol{\beta} \rangle + \frac{1}{2}(\boldsymbol{\beta} - \overline{\boldsymbol{\beta}})^T \{\nabla^2 L(\overline{\boldsymbol{\beta}}) - \nabla^2 L_j(\overline{\boldsymbol{\beta}})\}(\boldsymbol{\beta} - \overline{\boldsymbol{\beta}}),$$

for $j = 1, \dots K$. Here $\overline{\boldsymbol{\beta}} = \left(\sum_{j=1}^{K} \widehat{V}_j^{-1}\right)^{-1} \sum_{j=1}^{K} \widehat{V}_j^{-1} \widehat{\boldsymbol{\beta}}_j$ is an initial value that can be obtained using the inverse variance weighted average of estimates $\widehat{\boldsymbol{\beta}}_j$ by fitting local Cox models at each site[44], and the local log partial likelihood of the $j$-th site can be structured by $L_j(\boldsymbol{\beta}) = \frac{1}{n_j}\sum_{i=1}^{n_j} \delta_{ij} \log \frac{\exp(\boldsymbol{\beta}^T x_{ij})}{\sum_{s \in R_j(T_{ij})} \exp(\boldsymbol{\beta}^T x_{ij})}$. The first and second gradients of the surrogate likelihood function is calculated by weighting $\nabla L_j(\overline{\boldsymbol{\beta}})$ and $\nabla^2 L_j(\overline{\boldsymbol{\beta}})$ across all sites: $\nabla L(\overline{\boldsymbol{\beta}}) = \frac{1}{N}\sum_{j=1}^{K} n_j \nabla L_j(\overline{\boldsymbol{\beta}})$ and $\nabla^2 L(\overline{\boldsymbol{\beta}}) = \frac{1}{N}\sum_{j=1}^{K} n_j \nabla^2 L_j(\overline{\boldsymbol{\beta}})$, with the details for calculating $\nabla L_j(\overline{\boldsymbol{\beta}})$ and $\nabla^2 L_j(\overline{\boldsymbol{\beta}})$ available in Duan et al.[44] and Luo et al.[48].

Next, the global estimator of $\boldsymbol{\beta}_j$ is approximated by optimizing the above surrogate likelihood function. Then the final estimator $\widehat{\boldsymbol{\beta}}$ can be obtained by calculating inverse variance weighted average of all the $\widehat{\boldsymbol{\beta}}_j$s. This process for constructing the global model is one-shot[44], as illustrated in Figure 1. Since none of the shared files contains any patient level information but summary statistics, privacy is guaranteed. The final global survival scores are obtained by rounding coefficients of the global Cox model into integers and such that the range of total score is within the interval $[0, S_{max}]$, where $S_{max}$ is the maximum allowed score pre-decided with default value 100.

**4) Federated Model Selection and 5) Federated Model Evaluation**

Model selection is performed using parsimony plots generated on validation data, with variables added incrementally based on the variable ranking for the x-axis and integrated values of area under the receiver operating curves (iAUC) for the y-axis. Following Li et al.[28], we use a general model selection criteria defined by maximizing $\Psi_m = \sum w_j \phi_j(x_1, x_2, \ldots x_m)$, where $\phi_j$ measures a score's performance on the $j$-th validation set (e.g. iAUC value) and $m$ is a pre-specified number of total variables to include, uniform across all sites. Different constraints can be added for optimizing $\Psi_m$. For example, the total number of variables $m$ may not exceed an integer number $D$ for consideration of model parsimony. The set of covariates $\{x_1, x_2, \ldots x_m\}$ may also be set to satisfy certain subjective standards required by users' domain knowledge instead of strictly following the results of variable important analysis. Another option, which is a common practice, is to have $\Psi$ maximized by using a number of $d$ variables that is smaller than $m$, as long as increasing the number of variables from $d$ to $m$ has little impact on the change in $\Psi$. After final variables are confirmed, a new model is refitted via modules 2) and 3). The performance of this final model is validated on each site using their own training data, using iAUC for overall performance and AUC(t) values for performances for a given $t$-day survival.

The FedScore-Surv framework has been implemented in R 4.2.1 and the code is available at https://github.com/nliulab/FedScore.

**3.2 Experiment**

We conducted our experiments using two real-world heterogenous emergency department (ED) datasets: MIMIC-IV-ED[49] and the electronic health records (EHR) from Singapore General Hospital (SGH) data[50]. The EHR data of SGH was extracted from the SingHealth Electronic Health Intelligence System. A waiver of consent was granted for EHR data collection and retrospective analysis, and the study has been approved by the Singapore Health Services' Centralized Institutional Review Board, with all data deidentified.

The time-to-event outcome in this study is 30-day inpatient mortality after ED admission. The candidate variables include age, gender, pulse (beats/min), respiration (times/min), oxygen saturation (%), diastolic blood pressure (mm Hg), systolic blood pressure (mm Hg), and comorbidities including myocardial infarction, congestive heart failure, peripheral vascular disease, peptic ulcer disease, stroke, dementia, chronic pulmonary disease, hemiplegia or paraplegia, kidney disease, liver disease, diabetes and connective tissue disease. We follow the data extraction pipelines by Xie et al.[12] to pre-process the MIMIC-IV-ED dataset. This resulted in the creation of a master dataset, from which we formed a study population of 7177 samples by filtering only ED admissions of Asian patients aged 21 and older and removing observations with missing values. Similarly, for the SGH dataset, we obtained a study population with a total sample size of 43,408 by filtering the original SGH dataset for ED admissions of adult Asian patients in 2020 after excluding observations that have missingness.

For demonstration purposes, we randomly divided the MIMIC study cohort evenly into 2 sites and the SGH study cohort into 4 sites in the proportion of 10%, 20%, 30%, and 40%, leading to a total of 6 sites in our study. To predict the time-to-event outcome, we first employed our proposed framework and obtained a federated score trained using all six sites without sharing data across sites. For baseline comparison, we also employed AutoScore-Survival[18], a pre-existing tool for generating time-to-event clinical scores, to independently create local models for each site. To ensure fair comparisons between federated scores and local scores, we selected all models based on the corresponding parsimony plots, adhering to a predetermined criterion that the maximum allowable number of variables in a model should not exceed 10. In particular, during the model selection process, additional variables were included only if they contributed to a significant improvement in the integrated AUC value, as discussed in section 2.1. We used the default cutoff and weighting options as specified in Section 2.1 during the development of the survival scoring system and did not include expert knowledge refinement for a straightforward demonstration purposes.

## 4. Results

The final MIMIC cohort and SGH cohort obtained for this study have sample size 7177 and 43,408, respectively. As depicted in Figure 2, a total of 6 sites were formed, with a total sample size ranging from 3546 to 17,471. A comprehensive summary of the baseline characteristics of cohorts of each participating site can be found in eTable 1, where heterogeneity in data distribution can be observed. All testing and training sets were obtained at a ratio of 4:6, and 5-fold cross validation was employed for validation purpose during the parsimony steps of model development. The log rank test conducted across all Kaplan-Meier (KM) curves from all six sites yielded a p-value of 0.010, indicating significance of heterogeneity among the KM curves. Figure 3 depicts the KM curves and the corresponding number of patients at risk at each time point for both the training and testing datasets.

We compared the performance of the FedScore-Surv model with local scores on the testing data of each site. In Figure 4, the AUC(t) values and their 95% confidence intervals (CIs) of federated and local model are plotted against the change of time t, from day 1 to day 30, where each subplot represents the results on the testing data of a local site. The AUC(t) curve for the federated model is in green, and the one for local score model is in blue, with shadows of corresponding colours representing the 95% CIs of AUC(t). The iAUC values are also plotted as horizontal lines in each subplot. eTable 2 of Supplementary further reports all model's iAUC values and corresponding CIs on each site. The scoring tables for federated model and each local model can be found in eTable 3, and the corresponding parsimony plots are available in eFigure 1 of the Supplementary Materials.

Three main observations can be summarized based on the information presented in Figure 3 and eTable 2: 1) the federated model brings benefits to all local sites in terms of having an iAUC value higher than those of local models; 2) the federated model brings benefits to all local sites in terms of improving the AUC(t), in particular for later time points; 3) the federated model is more efficient than local models given that the CIs of AUC (t) at most time points are narrower than those of local models.

## 5. Discussion

FedScore-Surv is a communication-efficient framework with which users can conveniently collaborate in a privacy preserving way to create unified time-to-event scores across multiple sites with heterogeneous survival data. The framework is scalable and can be modified based on users' specific needs for clinical questions, allowing the engagement of domain expertise. The proposed method can offer potential solutions for improving both the accuracy and confidence of prediction models, bringing benefits to not only one but potentially all FL participants.

The prevalence of censored data in healthcare has catalysed the development and widespread application of survival analysis techniques. The approach is especially common in clinical studies with long-term follow-ups, encompassing various fields such as cancer research[51–53] and rare diseases[54–56]. Broadening our perspective beyond direct clinical fields to relevant domains like health economics, where models are expected to provide insights over individuals' lifetimes—often exceeding the available data span[57]—also underscores the continued necessity of survival analysis. Meanwhile, inaccurate model specification in such studies can result in very different decision-making, given that survival is often extrapolated beyond the data range[58]. Therefore, in cases where relevant long-term survival data are lacking, accurate modelling of short-term data becomes particularly important[58]. Consequently, leveraging information from other sites via FL could potentially enhance model confidence and bring benefits for participants, especially those facing challenges of inadequate cohort size and/or low data quality.

As shown in Figure 4, in our demonstrated example with ED data, sites 1-3 have relatively small sample sizes, and the federated score fitted via FedScore-Surv demonstrates a more significant improvement in the iAUC values compared to the baseline local scores. For sites 4-6 with larger sample sizes, although the improvement in iAUC is relatively less pronounced than for sites 1-3, our proposed method reduces the width of CIs for most AUC(t) values. This holds potential real-world clinical significance that, under the general expectation derived from common practices[40] of comparing FL models and local models in cross-silo FL applications in healthcare, all participating sites may derive more or less some benefits from such training processes. This, in turn, could enhance these sites' willingness to engage in real-world FL collaborations.

While we used partially artificially partitioned ED datasets for proof-of-concept, the example is sufficient to claim that our proposed method is capable of handling real-world heterogeneous survival data, supported by the fact that the MIMIC and SGH cohorts originate from two different countries. As shown in eTable 1 and eTable 2, the MIMIC and SGH cohorts differ not only in covariate distributions and outcome prevalence, but also in the test statistics of KM curves. This aligns with most real-world scenarios, where survival data collected across different sources can vary due to heterogeneous demographics and distinct clinical practices.

Our proposed method is communication-efficient and easy to implement. Participants only need to broadcast and receive summary statistics without patient-level information for one round of iteration. In contrast to engineering-based[40] FL frameworks, which usually require the coordination of a central server[41,59] and face data leakage issues[60,61], the implementation of our method can be easily achieved without dealing with potential limitations of participants' local data management systems when connecting to external servers, as discussed by Li et al.[62].

**Limitation and future work**

In this study, the results were obtained using artificially partitioned data derived from two real-world datasets, one of which is publicly available and the other locally accessible, instead of through a real-world collaboration between isolated regions or countries. In our forthcoming research, we intend to validate the proposed method by fostering local and international collaborations to model time-to-event outcomes of clinical interest and further examine the applicability of our method in real-world clinical decision-makings.

## 6. Conclusion

We have introduced FedScore-Surv, a privacy-preserving scoring system designed for time-to-event clinical outcomes and used a 30-day inpatient survival prediction task with real-world heterogeneous datasets for proof of concept. Our results clearly indicate the ability of FedScore-Surv to create robust federated time-to-event clinical scores, bring potential benefits to participating sites by offering federated models that outperform their locally developed counterparts. Our method is communication-efficient and can be easily implemented by users while avoiding any data leakage and privacy issues. While demonstrated with ED data, our method can be applied to a wide range of healthcare contexts to model time-to-event outcomes and bring benefits, especially to data sources with small and insufficient sample size.

# CRediT author statement

**Siqi Li:** Conceptualization, Data curation, Software, Formal analysis, Investigation, Methodology, Project administration, Writing – original draft, Writing – review & editing. **Yuqing Shang:** Data curation, Software, Formal analysis, Investigation, Methodology, Writing – original draft, Writing – review & editing. **Ziwen Wang:** Data curation, Formal analysis, Investigation, Methodology, Writing – original draft, Writing – review & editing **Qiming Wu:** Formal analysis, Investigation, Software, Writing – review & editing **Di Miao:** Formal analysis, Investigation, Writing – review & editing **Yilin Ning:** Data curation, Investigation, Writing – review & editing. **Chuan Hong:** Validation, Investigation, Writing – review & editing. **Marcus Eng Hock Ong:** Validation, Investigation, Writing – review & editing. **Bibhas Chakraborty:** Validation, Investigation, Writing – review & editing. **Nan Liu:** Conceptualization, Investigation, Methodology, Project administration, Funding acquisition, Resources, Supervision.

# Funding

This work was supported by the Duke/Duke-NUS Collaboration grant. The funder of the study had no role in study design, data collection, data analysis, data interpretation, or writing of the report.

# Competing Interests

NL, SL and MEHO hold a patent related to the federated scoring system. The other authors declare no competing interests.

# Code Availability

The R code used for this study is available at: https://github.com/nliulab/FedScore.

**Figure 1**. Flowchart of the FedScore-Surv framework.

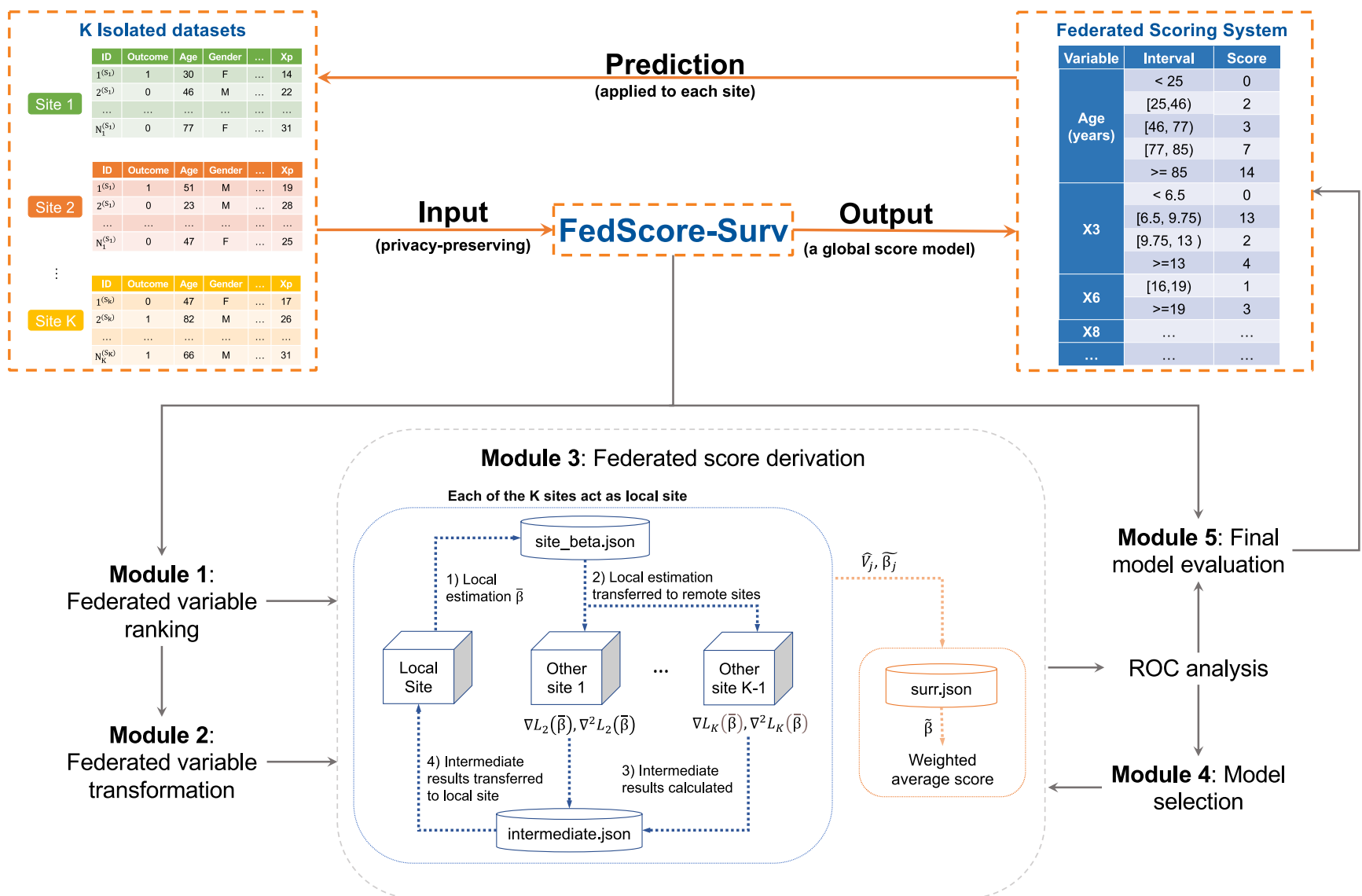

**Figure 2**. Flowchart of the study cohorts' formation. SGH: Singapore General Hospital

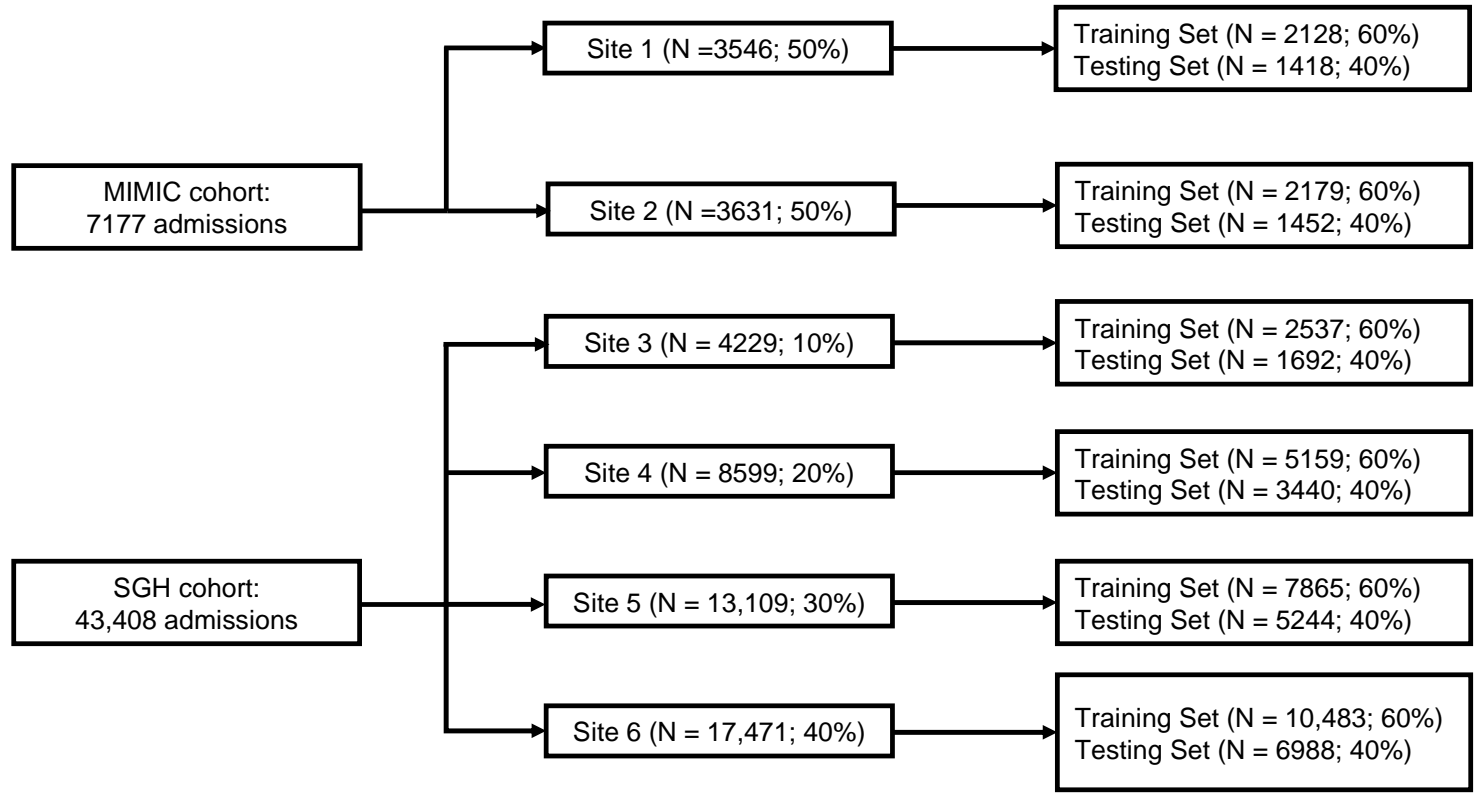

**Figure 3**. Kaplan-Meier curves and change of patients at risk in the testing sets of each local site.

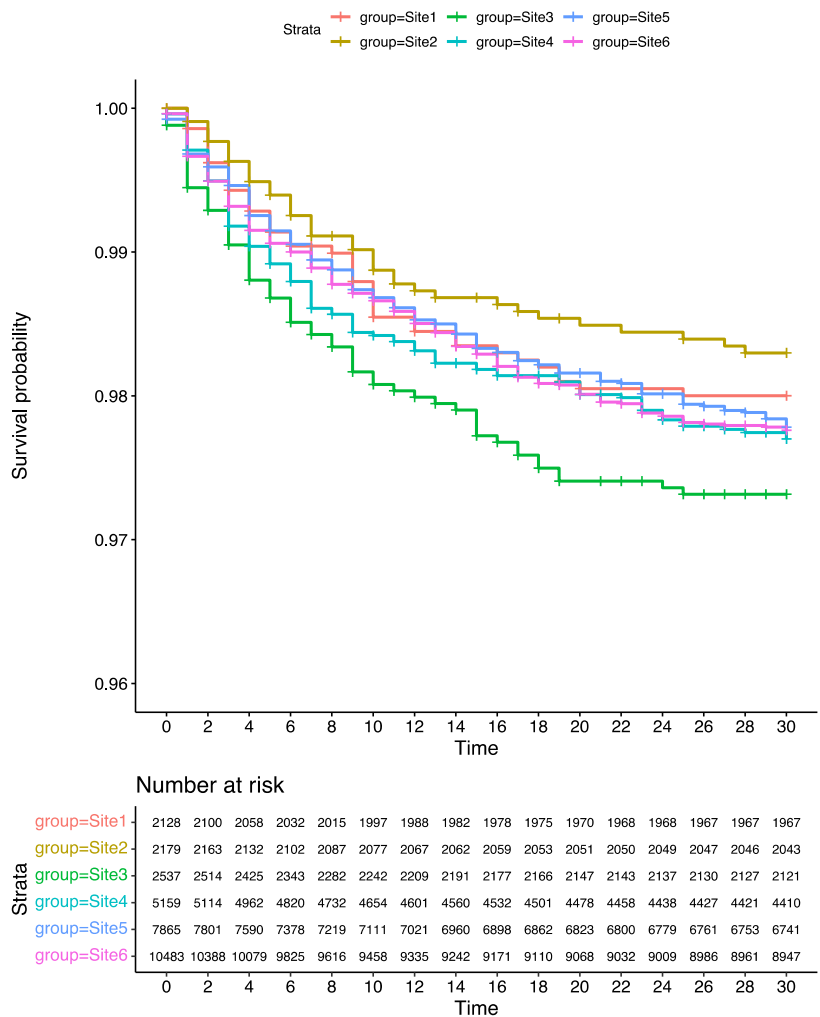

Training

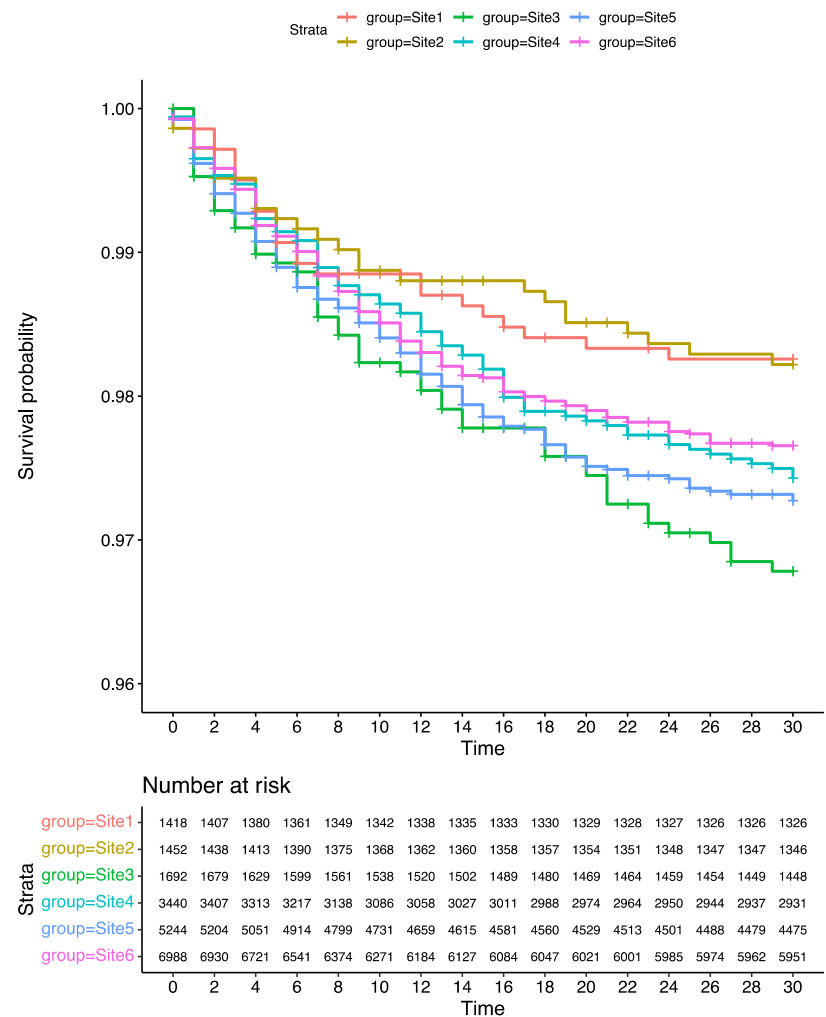

Testing

**Figure 4**. Performance comparison of FedScore and local scores based on AUC(t) and iAUC on each site.

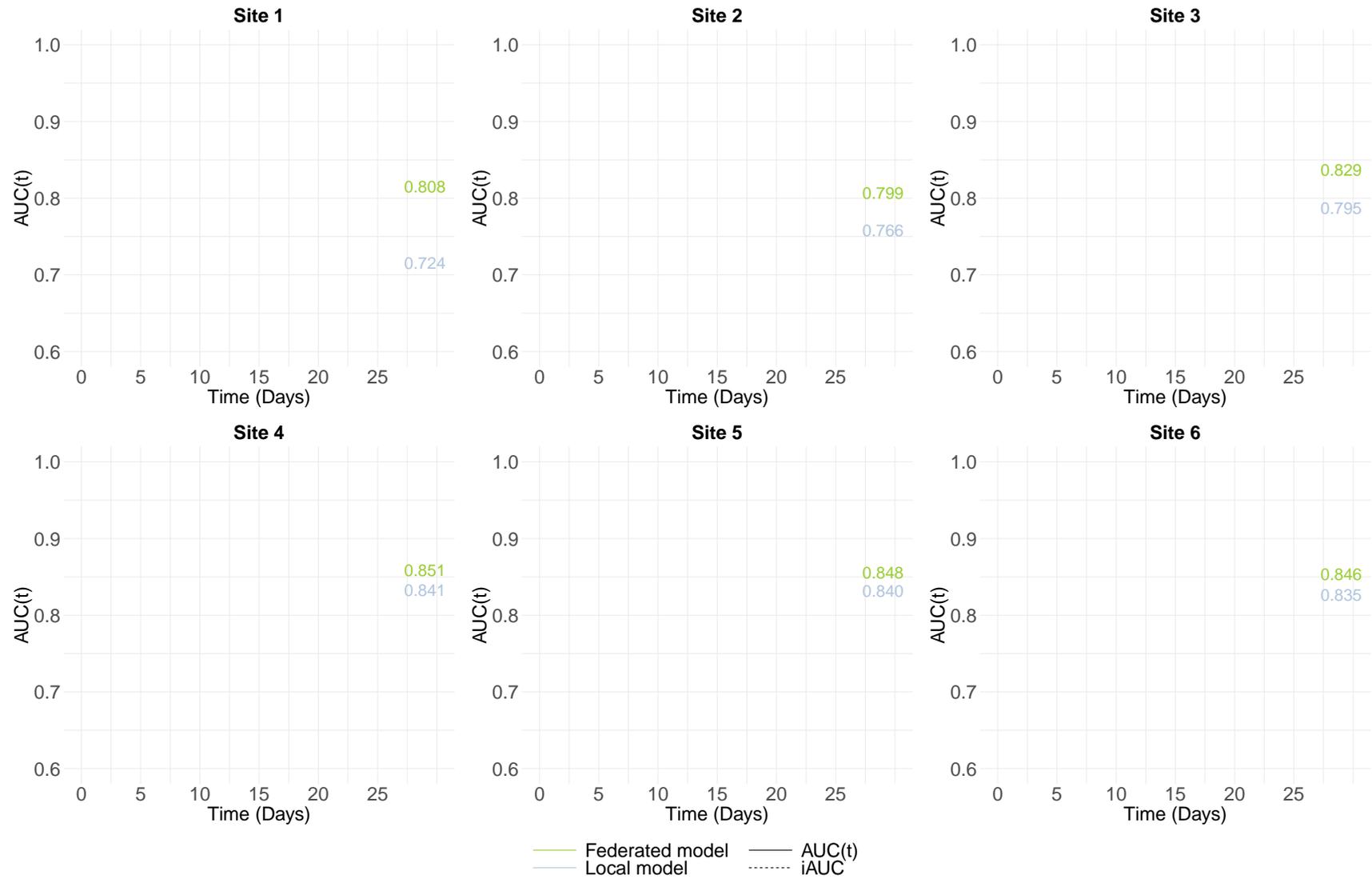

# Supplementary Materials

**eTable 1**: Description of the study cohorts.

|  | MIMIC | | SGH | | | |
|---|---|---|---|---|---|---|
|  | **Site 1** | **Site 2** | **Site 3** | **Site 4** | **Site 5** | **Site 6** |
| # Episodes | 3546 | 3631 | 4229 | 8599 | 13109 | 17471 |
| Survival time, days (mean (SD)) | 28.28 (6.36) | 28.46 (5.99) | 26.61 (8.28) | 26.83 (8.05) | 26.86 (8.00) | 26.79 (8.09) |
| 30 day in-patient mortality | 65 ( 1.8) | 61 ( 1.7) | 115 ( 2.7) | 194 ( 2.3) | 297 ( 2.3) | 373 ( 2.1) |
| Age, years (mean (SD)) | 57.19 (21.41) | 56.77 (21.82) | 62.60 (18.46) | 62.63 (18.62) | 62.87 (18.51) | 62.73 (18.59) |
| Vital Signs, mean (SD) | | | | | | |
| Pulse, bpm | 85.86 (19.14) | 87.09 (19.53) | 80.23 (18.17) | 80.14 (17.93) | 79.66 (17.87) | 80.10 (17.96) |
| Respiration, bpm | 17.66 (2.76) | 17.71 (2.67) | 17.30 (1.53) | 17.29 (1.50) | 17.26 (1.54) | 17.31 (1.52) |
| Oxygen saturation, % | 98.33 (2.68) | 98.22 (3.17) | 97.80 (4.55) | 97.79 (4.38) | 97.83 (4.21) | 97.76 (4.72) |
| Diastolic blood pressure, mmHg | 75.07 (22.10) | 74.95 (20.09) | 69.05 (15.60) | 68.91 (15.53) | 68.78 (15.32) | 68.82 (15.40) |
| Systolic blood pressure, mmHg | 132.77 (27.69) | 132.58 (23.99) | 128.22 (22.10) | 127.82 (21.90) | 128.01 (21.89) | 127.98 (21.74) |
| Comorbidities | | | | | | |
| Myocardial infarction | 149 ( 4.2) | 133 (3.7) | 260 ( 6.1) | 506 ( 5.9) | 760 ( 5.8) | 1040 ( 6.0) |
| Congestive heart failure | 342 ( 9.6) | 356 (9.8) | 354 ( 8.4) | 659 ( 7.7) | 965 ( 7.4) | 1343 ( 7.7) |
| Peripheral vascular disease | 160 ( 4.5) | 163 (4.5) | 191 ( 4.5) | 377 ( 4.4) | 599 ( 4.6) | 742 ( 4.2) |
| Stroke | 201 ( 5.7) | 194 (5.3) | 507 (12.0) | 1057 (12.3) | 1538 (11.7) | 2036 (11.7) |
| Dementia | 76 ( 2.1) | 71 (2.0) | 177 ( 4.2) | 351 ( 4.1) | 579 ( 4.4) | 738 ( 4.2) |
| Chronic pulmonary disease | 372 (10.5) | 422 (11.6) | 288 ( 6.8) | 596 ( 6.9) | 905 ( 6.9) | 1251 ( 7.2) |
| Rheumatoid disease | 72 ( 2.0) | 75 (2.1) | 69 ( 1.6) | 84 ( 1.0) | 188 ( 1.4) | 236 ( 1.4) |
| Peptic ulcer disease | 83 ( 2.3) | 106 (2.9) | 82 ( 1.9) | 212 ( 2.5) | 288 ( 2.2) | 356 ( 2.0) |
| Mild liver disease | 214 ( 6.0) | 231 (6.4) | 240 ( 5.7) | 479 ( 5.6) | 641 ( 4.9) | 830 ( 4.8) |
| Severe liver disease | 88 ( 2.5) | 82 (2.3) | 77 ( 1.8) | 194 ( 2.3) | 234 ( 1.8) | 326 ( 1.9) |
| Diabetes without chronic complications | 394 (11.1) | 359 (9.9) | 109 ( 2.6) | 228 ( 2.7) | 350 ( 2.7) | 494 ( 2.8) |
| Diabetes with complications | 264 ( 7.4) | 273 (7.5) | 1317 (31.1) | 2654 (30.9) | 4165 (31.8) | 5485 (31.4) |
| Hemiplegia or paraplegia | 41 ( 1.2) | 62 (1.7) | 263 ( 6.2) | 534 ( 6.2) | 743 ( 5.7) | 998 ( 5.7) |
| Kidney disease | 447 (12.6) | 471 (13.0) | 983 (23.2) | 1944 (22.6) | 3038 (23.2) | 3929 (22.5) |
| Gender (Male) | 1700 (47.9) | 1679 (46.2) | 2190 (51.8) | 4414 (51.3) | 6864 (52.4) | 9164 (52.5) |

[a]Data are presented as count (percentage) of patients unless otherwise indicated.

**eTable 2**: Comparison of performance of FedScore model with baseline models.

| Model | Number of Variables | Testing Data | | | | | | | | | | | | Mean iAUC of each model on all 6 sites |
|---|---|---|---|---|---|---|---|---|---|---|---|---|---|---|
| | | MIMIC | | | | | | SGH | | | | | | |
| | | Site 1 | | Site 2 | | Site 3 | | Site 4 | | Site 5 | | Site 6 | | |
| | | iAUC | CI | iAUC | CI | iAUC | CI | iAUC | CI | iAUC | CI | iAUC | CI | |
| Model 1[a] | 10 | 0.724 | 0.592-0.816 | 0.715 | 0.598-0.824 | 0.715 | 0.599-0.864 | 0.722 | 0.613-0.851 | 0.716 | 0.591-0.859 | 0.715 | 0.608-0.839 | 0.718 |
| Model 2[b] | 4 | 0.763 | 0.656-0.857 | 0.766 | 0.626-0.852 | 0.748 | 0.645-0.837 | 0.749 | 0.623-0.886 | 0.763 | 0.638-0.860 | 0.755 | 0.641-0.841 | 0.757 |
| Model 3[c] | 6 | 0.799 | 0.742-0.853 | 0.801 | 0.745-0.854 | 0.795 | 0.730-0.849 | 0.791 | 0.741-0.851 | 0.795 | 0.724-0.856 | 0.788 | 0.725-0.850 | 0.795 |
| Model 4[d] | 10 | 0.847 | 0.797-0.894 | 0.846 | 0.800-0.886 | 0.843 | 0.790-0.888 | 0.841 | 0.775-0.889 | 0.843 | 0.792-0.891 | 0.840 | 0.800-0.879 | 0.843 |
| Model 5[e] | 10 | 0.841 | 0.806-0.868 | 0.84 | 0.812-0.871 | 0.839 | 0.805-0.873 | 0.841 | 0.802-0.875 | 0.840 | 0.806-0.874 | 0.840 | 0.805-0.873 | 0.840 |
| Model 6[f] | 8 | 0.835 | 0.806-0.864 | 0.834 | 0.799-0.864 | 0.831 | 0.797-0.861 | 0.835 | 0.801-0.862 | 0.834 | 0.802-0.865 | 0.835 | 0.803-0.868 | 0.834 |
| Model federated[g] | 9 | 0.808 | 0.715-0.890 | 0.799 | 0.718-0.894 | 0.829 | 0.759-0.900 | 0.851 | 0.778-0.903 | 0.848 | 0.808-0.887 | 0.846 | 0.802-0.875 | 0.830 |
| **Average iAUC of all 6 local models on each site** | | 0.802 | | 0.800 | | 0.795 | | 0.797 | | 0.799 | | 0.796 | | |

[a-f] Local model obtained via AutoScore-Survival independently on site 1-6
[g] Federated model obtained via FedScore-Surv

**eTable 3**: Scoring tables.
(a) Scoring table of local model generated on site 1 via AutoScore-Survival.

| Variable | Interval | Point |
| --- | --- | --- |
| Pulse rate (per minute) | <70 | 6 |
| | [70,80) | 0 |
| | [80,89) | 9 |
| | [89,101) | 7 |
| | >=101 | 15 |
| Systolic blood pressure (mmHg) | <113 | 9 |
| | [113,124) | 4 |
| | [124,136) | 3 |
| | [136,152) | 0 |
| | >=152 | 1 |
| Age (years) | <33 | 0 |
| | [33,54) | 7 |
| | [54,66) | 9 |
| | [66,77) | 3 |
| | >=77 | 18 |
| Diastolic blood pressure (mmHg) | <63 | 4 |
| | [63, 71) | 6 |
| | [71,78) | 0 |
| | [78,86) | 7 |
| | >=86 | 6 |
| Oxygen saturation (%) | <97 | 21 |
| | [97,98) | 0 |
| | >=98 | 12 |
| Respiration rate (per minute) | <16 | 6 |
| | [16,18) | 0 |
| | >=18 | 4 |
| Rheumatoid disease | Yes | 1 |
| | No | 0 |
| Severe liver disease | Yes | 0 |
| | No | 4 |
| Chronic pulmonary disease | Yes | 10 |
| | No | 0 |
| Gender | Female | 0 |
| | Male | 7 |

(b) Scoring table of local model generated on site 2 via AutoScore-Survival.

| Variable | Interval | Point |
| --- | --- | --- |
| Systolic blood pressure (mmHg) | <113 | 4 |
| | [113,124) | 3 |
| | >=124 | 0 |
| Age (years) | <32 | 0 |

| | [32,54) | 44 |
| | [54,78) | 46 |
| | >=78 | 48 |
| Oxygen saturation (%) | <97 | 3 |
| | [97,99) | 0 |
| | >=99 | 1 |
| Respiration rate (per minute) | <16 | 0 |
| | >=16 | 44 |

(c) Scoring table of local model generated on site 3 via AutoScore-Survival.

| **Variable** | **Interval** | **Point** |
| --- | --- | --- |
| Pulse rate (per minute) | <64 | 0 |
| | [64,72) | 1 |
| | [72,81) | 15 |
| | [81,100) | 5 |
| | >=100 | 14 |
| Systolic blood pressure (mmHg) | <108 | 19 |
| | [108,119) | 18 |
| | [119,130) | 8 |
| | [130,145) | 3 |
| | >=145 | 0 |
| Age (years) | <46 | 0 |
| | [46,70) | 22 |
| | [70,80) | 24 |
| | >=80 | 37 |
| Diastolic blood pressure (mmHg) | <58 | 14 |
| | [58, 63) | 7 |
| | [63,70) | 0 |
| | [70,79) | 5 |
| | >=79 | 6 |
| Oxygen saturation (%) | <96 | 7 |
| | [96,98) | 3 |
| | >=98 | 0 |
| Respiration rate (per minute) | <16 | 7 |
| | [16,17) | 1 |
| | [17,18) | 0 |
| | >=18 | 9 |

(d) Scoring table of local model generated on site 4 via AutoScore-Survival.

| **Variable** | **Interval** | **Point** |
| --- | --- | --- |
| Pulse rate (per minute) | <65 | 1 |
| | [65,72) | 2 |
| | [72,81) | 0 |

| | [81,100) | 9 |
| | >=100 | 15 |
| Systolic blood pressure (mmHg) | <109 | 12 |
| | [109,120) | 3 |
| | [120,130) | 2 |
| | >=130 | 0 |
| Age (years) | <46 | 0 |
| | [46,61) | 12 |
| | [61,70) | 19 |
| | [70,79) | 21 |
| | >=79 | 28 |
| Diastolic blood pressure (mmHg) | <58 | 6 |
| | [58, 64) | 0 |
| | [64,70) | 1 |
| | [70,79) | 2 |
| | >=79 | 5 |
| Oxygen saturation (%) | <96 | 5 |
| | [96,98) | 0 |
| | >=98 | 2 |
| Respiration rate (per minute) | <16 | 4 |
| | [16,17) | 1 |
| | [17,18) | 0 |
| | >=18 | 2 |
| Myocardial infarction | Yes | 15 |
| | No | 0 |
| Severe liver disease | Yes | 10 |
| | No | 0 |
| Congestive heart failure | Yes | 2 |
| | No | 0 |
| Peptic ulcer disease | Yes | 0 |
| | No | 2 |

(e) Scoring table of local model generated on site 5 via AutoScore-Survival.

| **Variable** | **Interval** | **Point** |
| --- | --- | --- |
| Pulse rate (per minute) | <80 | 0 |
| | [80,100) | 6 |
| | >=100 | 13 |
| Systolic blood pressure (mmHg) | <109 | 16 |
| | [109,119) | 11 |
| | [119,145) | 4 |
| | >=145 | 0 |
| Age (years) | <46 | 0 |
| | [46,61) | 20 |
| | [61,70) | 19 |

| Variable | Interval | Point |
| --- | --- | --- |
| | [70,79) | 24 |
| | >=79 | 32 |
| Diastolic blood pressure (mmHg) | <57 | 4 |
| | [57, 64) | 1 |
| | [64,70) | 0 |
| | [70,78) | 2 |
| | >=78 | 6 |
| Oxygen saturation (%) | <96 | 4 |
| | [96,98) | 0 |
| | >=98 | 2 |
| Respiration rate (per minute) | <16 | 7 |
| | [16,17) | 1 |
| | [17,18) | 0 |
| | >=18 | 6 |
| Myocardial infarction | Yes | 13 |
| | No | 0 |
| Stroke | Yes | 3 |
| | No | 0 |
| Congestive heart failure | Yes | 1 |
| | No | 0 |
| Peptic ulcer disease | Yes | 5 |
| | No | 0 |

(f) Scoring table of local model generated on site 6 via AutoScore-Survival.

| Variable | Interval | Point |
| --- | --- | --- |
| Pulse rate (per minute) | <64 | 2 |
| | [64,81) | 0 |
| | [81,100) | 5 |
| | >=100 | 12 |
| Systolic blood pressure (mmHg) | <109 | 15 |
| | [109,120) | 10 |
| | [120,131) | 3 |
| | [131,145) | 5 |
| | >=145 | 0 |
| Age (years) | <46 | 0 |
| | [46,61) | 18 |
| | [61,70) | 23 |
| | [70,79) | 24 |
| | >=79 | 31 |
| Diastolic blood pressure (mmHg) | <58 | 5 |
| | [58, 64) | 1 |
| | [64,70) | 0 |
| | >=70 | 4 |
| Oxygen saturation (%) | <96 | 6 |

| | [96,98) | 0 |
| | >=98 | 6 |
| Respiration rate (per minute) | <16 | 6 |
| | [16,17) | 0 |
| | [17,18) | 2 |
| | >=18 | 4 |
| Myocardial infarction | Yes | 11 |
| | No | 0 |
| Severe liver disease | Yes | 13 |
| | No | 0 |

(g) Scoring table of federated model generated via FedScore-Surv.

| Variable | Interval | Point |
| --- | --- | --- |
| Pulse rate (per minute) | <82 | 0 |
| | [82,100) | 4 |
| | >=100 | 12 |
| Systolic blood pressure (mmHg) | <109 | 15 |
| | [109,120) | 10 |
| | [120,131) | 3 |
| | [131,146) | 4 |
| | >=146 | 0 |
| Age (years) | <44 | 0 |
| | [44,60) | 18 |
| | [60,79) | 25 |
| | >=79 | 33 |
| Diastolic blood pressure (mmHg) | <58 | 5 |
| | [58, 65) | 1 |
| | [65,71) | 0 |
| | [71,79) | 3 |
| | >=79 | 4 |
| Oxygen saturation (%) | <96 | 6 |
| | [96,98) | 0 |
| | >=98 | 6 |
| Respiration rate (per minute) | <16 | 5 |
| | [16,18) | 0 |
| | >=18 | 4 |
| Myocardial infarction | Yes | 11 |
| | No | 0 |
| Congestive heart failure | Yes | 0 |
| | No | 1 |
| Severe liver disease | Yes | 12 |
| | No | 0 |

**eFigure 1**: Parsimony plots.

(a) - (f): Parsimony plots of local scoring models generated on site 1 to site 6 independently via AutoScore-Survival.

(g): Parsimony plot of federated scoring model generated via FedScore-Surv.

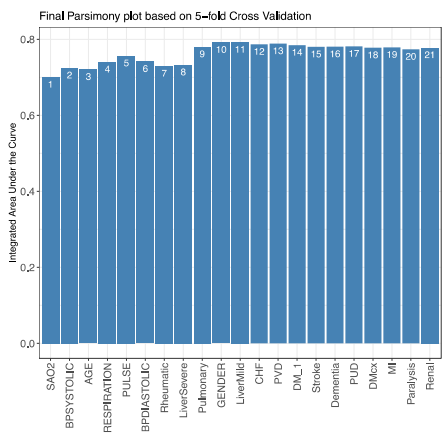
(a)

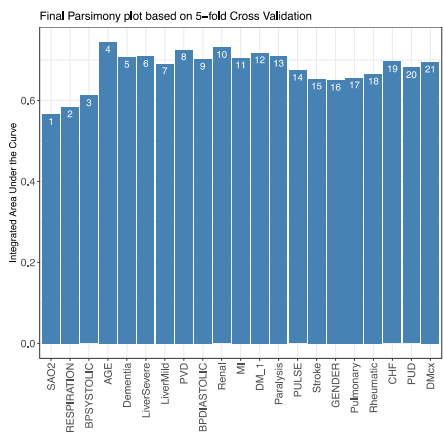
(b)

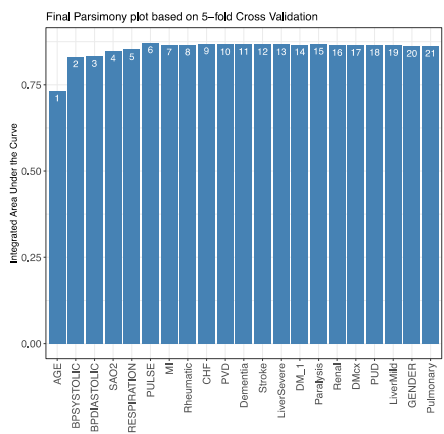
(c)

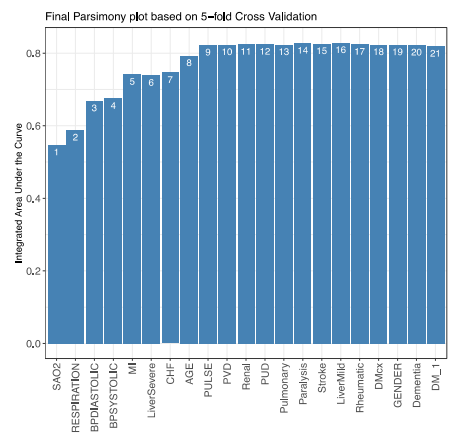
(d)

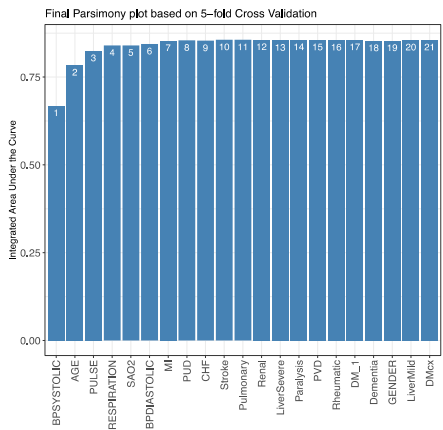
(e)

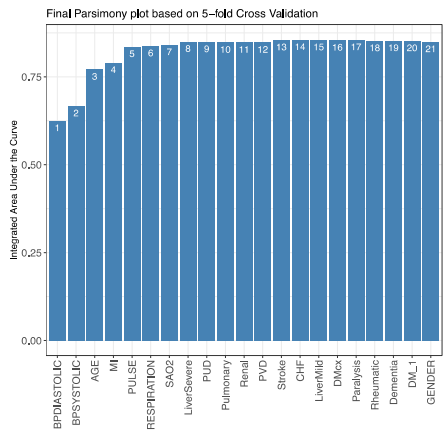
(f)

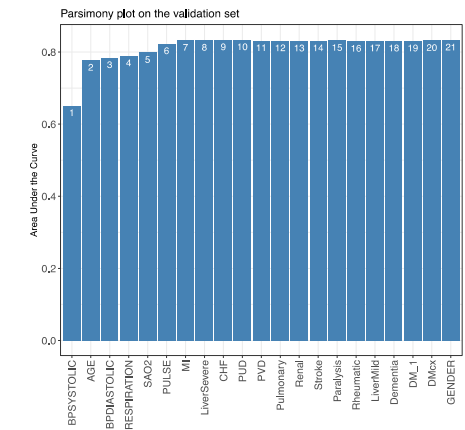
(g)

34